\documentclass{article}
\usepackage{spconf,amsmath,graphicx,hyperref}

\usepackage{graphicx}
\usepackage{amsmath, amssymb, amsfonts}
\usepackage{booktabs}
\usepackage{tabularx}

\usepackage{hyperref}
\hypersetup{colorlinks=true, linkcolor=blue, filecolor=magenta, urlcolor=cyan}

\title{WiFi-GEN: High-Resolution Indoor Imaging from WiFi Signals Using Generative AI}
\name{Jianyang Shi$^{1,5\dagger}$, Bowen Zhang$^{2\dagger}$, Amartansh Dubey$^3$, Ross Murch$^4$ and Liwen Jing$^1$*\thanks{$^\dagger$ Jianyang Shi and Bowen Zhang contributes equally to this work, *Corresponding author: Liwen Jing (jinglw@pcl.ac.cn)}}
\address{$^1$ Pengcheng Laboratory, 
$^2$College of Big Data and Internet, Shenzhen Technology University, 
\\$^3$Department of Electrical Engineering, Indian Institute of Technology Delhi,
\\$^4$Department of Electronic and Computer Engineering, HKUST,\\ $^5$ School of Computer Science and Technology, Harbin Institute of Technology (Shenzhen)}
\begin{document}
%
\maketitle
\begin{abstract}
Indoor imaging is a critical task for robotics and internet-of-things. WiFi as an omnipresent signal is a promising candidate for carrying out passive imaging and synchronizing the up-to-date information to all connected devices. This is the first research work to consider WiFi indoor imaging as a multi-modal image generation task that converts the measured WiFi power into a high-resolution indoor image.

Our proposed WiFi-GEN network achieves a shape reconstruction accuracy that is 275\% of that achieved by physical model-based inversion methods.
Additionally, the Fr\'{e}chet Inception Distance score has been significantly reduced by 82\%.
To examine the effectiveness of models for this task, the first large-scale dataset is released containing 80,000 pairs of WiFi signal and imaging target.
Our model absorbs challenges for the model-based methods including the non-linearity, ill-posedness and non-certainty into massive parameters of our generative AI network. 
The network is also designed to best fit measured WiFi signals and the desired imaging output. 
Code: https://github.com/CNFightingSjy/WiFiGEN
\end{abstract}
\section{Introduction and Related Work}
\label{sec:intro}

Indoor environment perception is increasingly important for robots, smart devices, and intelligent systems in homes, offices, and industrial settings.  WiFi-based indoor imaging and localization have attracted attention as cost-effective, privacy-preserving, and device-free solutions \cite{wang2025wicg}\cite{adib2013see}. Unlike image sensors, WiFi signals capture only limited target features (e.g., location, shape, movement) \cite{gao2024autosen}\cite{deshmukh2022physics}\cite{lyons2024wifiact}, inherently safeguarding privacy without additional computation. As a passive method, it requires no extra devices and minimal modification of WiFi access points.
WiFi imaging transforms WiFi signal measurements into images.

A promising approach to WiFi imaging is to formulate it as an inverse scattering problem (ISP), which has demonstrated superior performance over radio tomographic imaging (RTI) \cite{wilson2010radio}. WiFi ISP can be defined in two forms: full-wave, which requires both amplitude and phase information, and phaseless, which relies only on signal strength \cite{chen2020review}\cite{sanghvi2019embedding}\cite{jing2018detecting}. Although full-wave measurements are expensive to obtain in practice \cite{xu2024wicamera}, recent studies have shown that phaseless imaging based solely on power measurements is feasible \cite{xu2020deep}\cite{dubey2021enhanced}. Existing methods, however, still depend on approximate solutions to wave physics and struggle with the non-linear, ill-posed nature of the problem. Deep learning–based neural networks have been introduced to mitigate these challenges and improve reconstruction quality \cite{deshmukh2022unrolled}, yet the resulting images often fail to recover hollow structures.
In this work, we present a novel end-to-end generative AI framework for phaseless WiFi imaging, designed to reconstruct accurate, high-resolution images from limited measurements. We formulate the task as multi-modal image generation, inspired by models such as DALL-E \cite{rombach2022high}, where a dense input power matrix is transformed into a sparser output image matrix.

The main contributions of this work are as follows:
(1) WiFi-GEN is introduced as the first generative AI framework that formulates indoor WiFi imaging as a multi-modal image generation problem.
(2) A large-scale dataset of 80,000 WiFi–image pairs is constructed to facilitate data-driven solutions for this emerging task.
(3) Experimental results show that WiFi-GEN outperforms physics-based inversion methods, achieving more accurate shape reconstruction, higher-resolution imaging, and requiring fewer WiFi nodes.
\section{Problem Formulation and Method}\label{method}

\subsection{Problem Formulation}
Consider a typical indoor environment with a domain of interest $D\subset \mathbb{R} ^{2} $ with rectangular shape of $X \times Y$ m$^2$. WiFi nodes are placed at its boundary. In order to obtain measurements for phaseless inverse scattering problem (PD-ISP), each WiFi node needs to measure the signal power transmitted from all other nodes separately. Current WiFi devices can be configured to work as both transmitter and receiver, so the measurement matrix $\mathbf{W} $ obtained from $M$ nodes has the largest dimension of $M \times M-1$. In practice, all $M$ nodes take turns to act as the transmitter. While one of the nodes transmits, all other $M-1$ nodes measure the received power coming from a constant transmitting power level. This configuration is similar to conventional works in this research field \cite{dubey2021enhanced}. 

The PD-ISP aims to identify the material parameter (e.g. permittivity) in each cell of the discretized 2-D DoI of $\mathbf{I} 
(N_x \times N_y$), where $N_x = X/\Delta x$ and $N_y = Y/\Delta y$. For better resolution, $N_x \times N_y$ is normally much larger than $M \times M-1$, and the increase of $N_x$ and $N_y$ can provide more details in the DoI. For imaging task, detecting the shape is the priority, so in this work we only reconstruct the object shape rather than the material permittivity. 
Consequently, the final task is generating the $N_x \times N_y$ binary matrix $\mathbf{I}$ from the measured input matrix $\mathbf{W}$($M \times M-1 $). We consider the PD-ISP as a new cross-modal image generation problem and design a generative model called WIFI-GEN to visualize indoor objects through WiFi signals. 

\subsection{WiFi-GEN}

\begin{figure}
	\centering
	\includegraphics[scale=0.32]{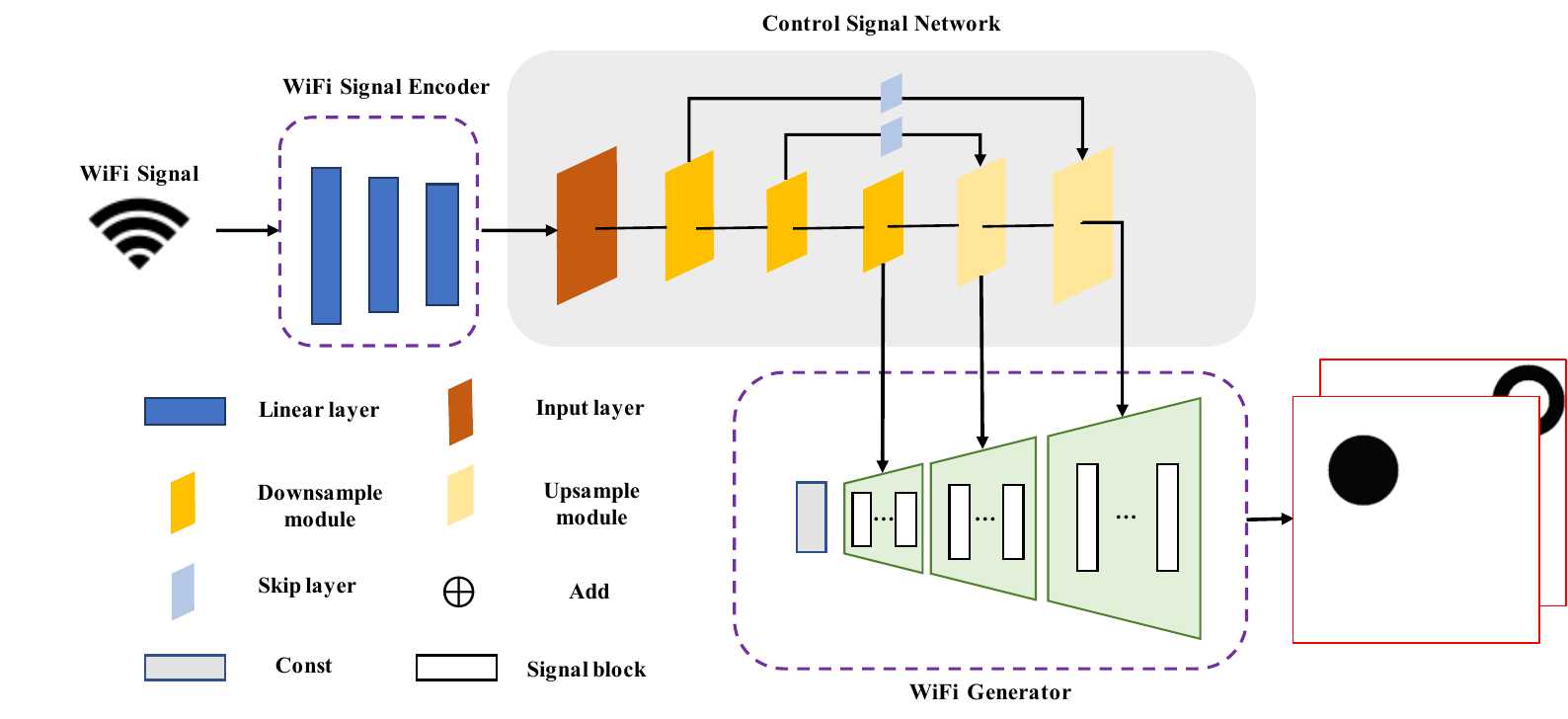}
	\caption{Framework of our proposed WiFi-GEN, where Downsample, Upsample and Signal Block are illustrated in detail in Figure \ref{fig:details}. Both input layer and skip layer consist of simple fully convolutional network.}
	\label{fig:model}
\end{figure}

\begin{figure}
	\centering
	\includegraphics[scale=0.3]{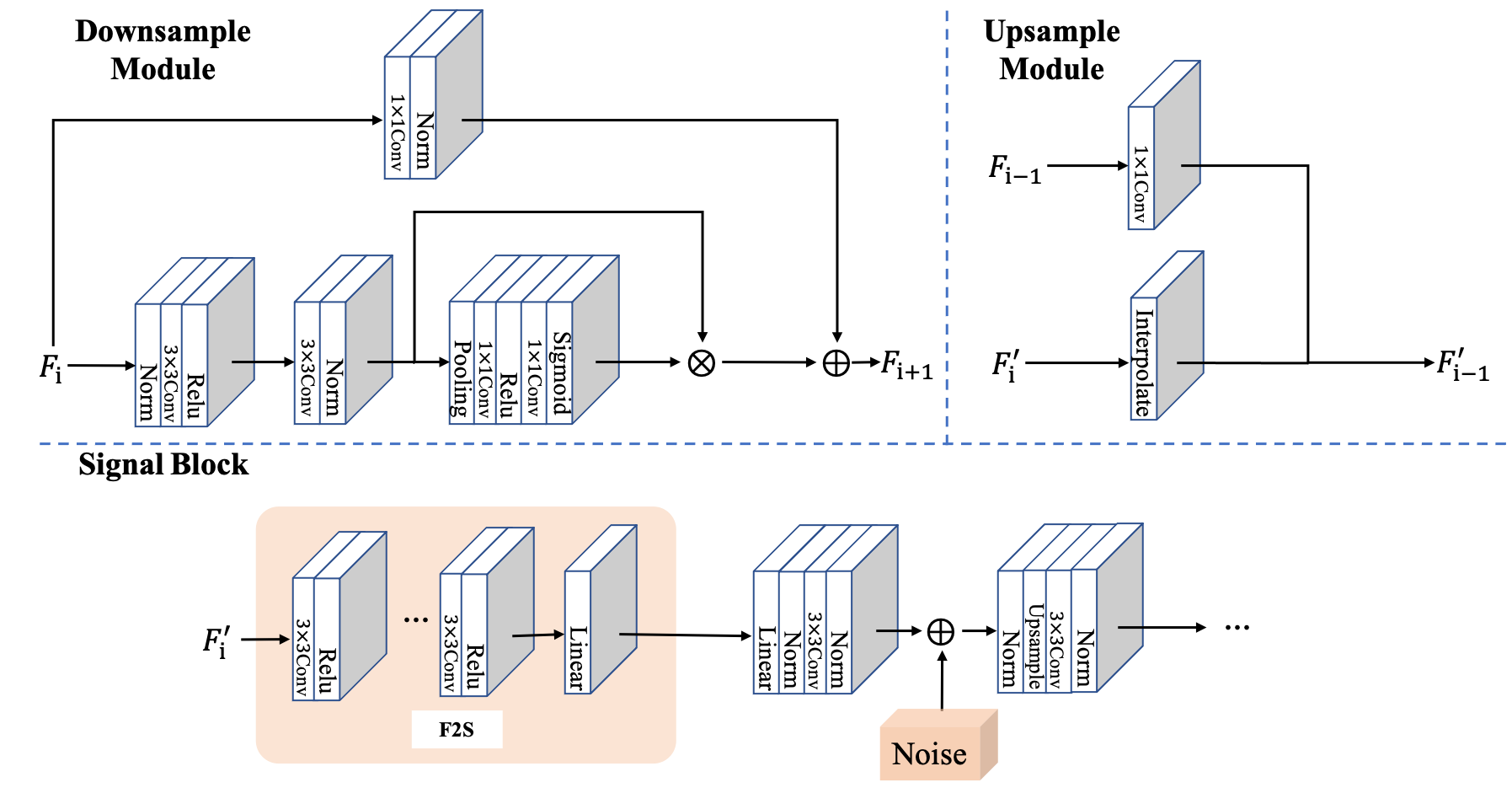}
	\caption{Structure details of Downsample, Upsample and Signal Block in WiFi-GEN. $F_i$ denotes the feature extracted by the i-th downsample module, $F_i^{'}$ denotes the feature extracted by the i-th upsample module.}
	\label{fig:details}
\end{figure}
Inspired by cross-modal image generation, this work proposes WiFi-GEN, an end-to-end network for WiFi-based indoor imaging. The model comprises three modules: (1) a WiFi signal encoder that maps raw signals into a latent space matrix via a learnable fully connected layer; (2) a hierarchical control network that extracts multi-level WiFi features and represents them as 512-dimensional vectors; and (3) a WiFi generator that integrates these features to guide high-quality image synthesis. The overall WiFi-GEN architecture is illustrated in Fig. \ref{fig:model}.

\textbf{WiFi Signal Encoder.} 
The encoder compresses WiFi power measurements into a compact latent space, removing redundancy from wave phenomena such as scattering and reflection. 
A three-layer fully connected network flattens the $19\times20$ signal into a 380-dimensional vector while preserving node-wise ordering. 
The output is embedded into a $16\times16$ latent matrix for subsequent feature extraction.  

\textbf{Control Signal Network.} 
The control network extracts multi-scale features from the latent space to guide accurate image generation. 
Inspired by UNet-like feature pyramids, it employs residual-based downsampling and convolutional upsampling to capture object variations in shape and size. 
Feature maps from selected layers (4, 21, 24) of the downsample path are forwarded to the generator, as illustrated in Fig.~\ref{fig:details}.  

\textbf{WiFi Generator.} 
The generator synthesizes the final image using hierarchical features from the control network. 
Following the StyleGAN paradigm, it integrates multi-scale feature vectors through signal blocks with varying convolution depths (6 for shallow, 5 for intermediate, 4 for deep features). 
These are progressively upsampled in the image perception module, with injected noise enhancing sensitivity to small objects. 
The detailed generator structure is shown in Fig.~\ref{fig:details}.

\subsection{Loss Function} 
In order to ensure the accuracy of the generated image shape, we introduce $L_2$ Loss to provide pixel-level generation supervision, with the specific form as follows,
\begin{equation}
	L_{2}(\mathbf{W}) = \left \| \mathbf{I}_{gen}-\rm{WiFi\text{-}GEN}(\mathbf{W}))  \right \| _{2} 
\end{equation}
where $\mathbf{W}$ represents the measurement matrix of WiFi power, and $\mathbf{I}_{gen}$ denotes the ground truth image of the indoor environment. In order to further measure the difference between two images and maintain the quality of imaging during training, we introduce LPIPS loss, which is defined as follows,
\begin{equation}
	L_{\mathrm{LPIPS} }(\mathbf{W} ) = \left \| F(\mathbf{I} _{gen})-F(\rm{WiFi\text{-}GEN}(\mathbf{W} ))  \right \| _{2} 
\end{equation}
where $F$ is a pre-trained perceptual feature extractor.

The overall objective function of WiFi-GEN is calculated as a weighted sum of the above two loss functions as follows:
\begin{equation}
	\label{lobj}
	\begin{split}
		\mathcal{L}(\mathbf{W})=\mathcal{L}_{2}+\gamma\mathcal{L}_{LPIPS}
	\end{split}
\end{equation}
where $\gamma$ is a hyper-parameter. 



\section{Simulation and Experimental Results}\label{results}
\subsection{Experimental Setup and Parameter Settings}
\textbf{Setup}:
To evaluate the proposed method, experiments were conducted in a $3 \times 3$ m$^2$ indoor area equipped with 20 WiFi nodes operating at 2.4\,GHz. 
The nodes were evenly placed along the room perimeter; one acted as the transmitter while the remaining 19 served as receivers, producing a $19 \times 20$ signal strength matrix per measurement. 
The algorithm aims to reconstruct the entire area (Domain of Interest) as a $256 \times 256$ binary image, delineating object contours within the space.


We used a simulation program to generate datasets for training and evaluating our generative models, creating image-WiFi measurement pairs for diverse scenario analysis. These results are in section \ref{sec:simulation}. Additionally, real-world tests with 20 ADALM-PLUTO radios in a lab setting, akin to the study in \cite{deshmukh2022physics} but with fewer WiFi nodes, further validated our method. Details on these physical experiments are in section \ref{sec:measurements}.


\textbf{Datasets:} 
Training generative models requires large-scale data, yet collecting WiFi measurements in physical environments is prohibitively labor-intensive. 
To address this, we constructed a simulated dataset by translating imaging results into synthetic WiFi signals. 
Four object shapes (circles, squares, triangles, and rings) with random sizes and positions were generated, producing 20,000 samples per shape. 
Each simulated result was converted into a binary image via threshold segmentation, with objects in black and background in white. 
This approach enables efficient data generation while ensuring diversity and coverage for robust training. 
We use 80\% of the data for training and 20\% for evaluation.

\textbf{Parameter Setting:} 
The input to the model is a $19 \times 20$ WiFi signal matrix, and the output is a $256 \times 256$ pixel image. 
Training is performed with a learning rate of 0.0001, batch size of 8, and hyper-parameter $\gamma=0.8$. 
All models are trained on an NVIDIA RTX A6000 GPU (48 GB memory), requiring 7 days to complete 50 epochs.  

\textbf{Evaluation Metrics:} 
Model performance is evaluated using Fr\'{e}chet Inception Distance (FID) \cite{heusel2017gans} and Intersection over Union (IoU). 
FID measures generative quality, while IoU evaluates the accuracy of object shape and position. 
IoU is defined as
\begin{equation}
	IoU = \frac{\mathbf{I}_{gen} \cap \mathbf{I}}{\mathbf{I}_{gen} \cup \mathbf{I}},
\end{equation}
where the numerator represents the overlapping area between generated and ground-truth boundaries, and the denominator represents their union.

\begin{figure}
	\centering
	\includegraphics[width=0.45\textwidth]{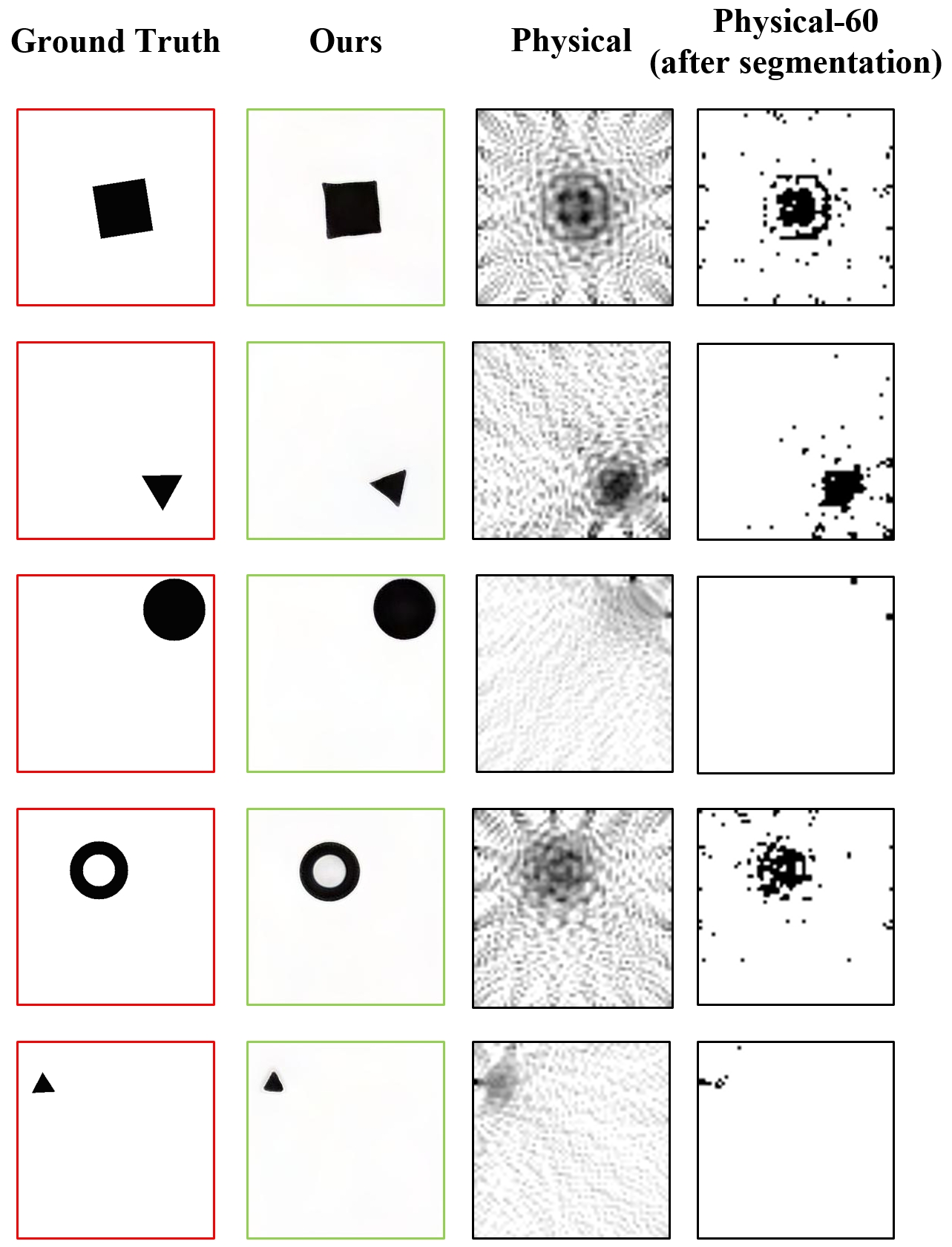}
	\caption{Performance of selected cases of our model in comparison with physical model-based methods. Our model demonstrates substantial improvement in shape reconstruction accuracy.}
	\label{fig:compare}
\end{figure}

\begin{table}[htbp]
	\centering
	\begin{tabular}{lcc}
		\toprule
		Model & IoU $\uparrow$   & FID $\downarrow$\\
		\midrule
		Physical &    -   & 436.752 \\
		Physical-60 & 0.289 & - \\
		Physical-256 & 0.305 & 270.486 \\
		Ours  & \textbf{0.795} & \textbf{79.675} \\
		\bottomrule
	\end{tabular}
	\linebreak 
	\caption{IoU and FID results of our proposed model in comparison with physical model-based methods.}
	
	\label{tab:compare}%
\end{table}%

\begin{table}[htbp]
	\centering
	\begin{tabular}{lcccc}
		\toprule
		Model & Circle & Rectangle & Triangle & Ring \\
		\midrule
		Physical-60  & 0.287 & 0.392 & 0.278 & 0.201 \\
		Physical-256 & 0.3  & 0.416 & 0.296 & 0.21 \\
		Ours     & \textbf{0.841} & \textbf{0.836} & \textbf{0.777} & \textbf{0.728} \\
		\bottomrule
	\end{tabular}
	\linebreak 
	\caption{IoU scores for four distinct shapes. Our proposed approach demonstrates a substantial enhancement in performance across all categories, ranging from two to three times. Notably, the ring, which is the most challenging shape to perceive, exhibited a remarkable 3.62-fold accuracy comparing with physical methods.}
	\label{tab:addlabel}%
\end{table}%








\begin{table*}[]
	\centering
	\begin{tabular}{cccccc}
		\hline
		Reference                                               & Frequency (GHz) & DoI Size ($m^2$)           & Reletive Permittivity($\epsilon_r$) & WiFi Nodes   & Image Resolution \\ \hline
		Xu et al. \cite{xu2020deep}            & 0.4             & $2.67 \times 2.67$ & 1.1--1.8                            & 16 Tx, 32 Rx & $32 \times 32$   \\
		Dubey et al. \cite{dubey2022phaseless} & 2.4             & $3 \times 3$ or $5 \times 5$  & 1.1+0.1i to 77+7i                   & 40 TRx or 20 TRx    & $60 \times 60$   \\
		WiFi-GEN                                                & 2.4             & $3 \times 3$     & 4+0.4i, 77+7i                       & 20 TRx       & $256 \times 256$ \\ \hline
	\end{tabular}
	\linebreak 
	\caption{Performance Comparison with Published Work.}
	\label{tab:related work}
\end{table*}

\subsection{Comparison to Traditional Methods}\label{sec:simulation}
To evaluate WiFi-GEN, we compare its performance with state-of-the-art physical model-based methods. 
Representative results for different object shapes are shown in Fig.~\ref{fig:compare}, while quantitative comparisons are given in Tables~\ref{tab:compare} and \ref{tab:addlabel}. 
A total of 16,000 data samples (20\% of the dataset across four shapes) are used for evaluation. 
WiFi-GEN outperforms traditional methods in three aspects: 
(1) it improves shape reconstruction accuracy by 275\% (IoU) and reduces FID by 82\%, while avoiding severe noise in generated images; 
(2) it achieves 3.62$\times$ higher accuracy for non-convex ring shapes, which are particularly challenging for physical models; 
and (3) it maintains robust performance near scene boundaries, where physical models—typically assuming far-field conditions—struggle within one wavelength of WiFi nodes.  

\textbf{Post-processing of physical methods:} 
Results from physical models suffer from significant noise. 
To mitigate this, we applied threshold segmentation to highlight salient regions. 
``Physical'' refers to the raw results, ``Physical-60'' to segmented $60\times60$ images, and ``Physical-256'' to segmented results resampled to $256\times256$.  

\textbf{Shape reconstruction accuracy:} 
Unlike physical models limited to $60\times60$ resolution, WiFi-GEN directly generates perceptual-quality $256\times256$ images. 
Fig.~\ref{fig:compare} illustrates its ability to recover four distinct shapes, while Tables~\ref{tab:compare} and \ref{tab:addlabel} confirm superior IoU and FID scores.  

\textbf{Performance on challenging cases:} 
WiFi-GEN accurately reconstructs small or boundary-positioned objects, such as peripheral triangles, and shows clear improvements for non-convex ring shapes, demonstrating robustness beyond traditional approaches.  

\textbf{Comparison with published work:} 
Further comparisons with existing studies are provided in Table~\ref{tab:related work}, covering operating frequency, DoI size, object permittivity, number of WiFi nodes, and output resolution.

\subsection{Physical Experiment}\label{sec:measurements}
\begin{figure}
	\centering
	\includegraphics[width=0.41\textwidth]{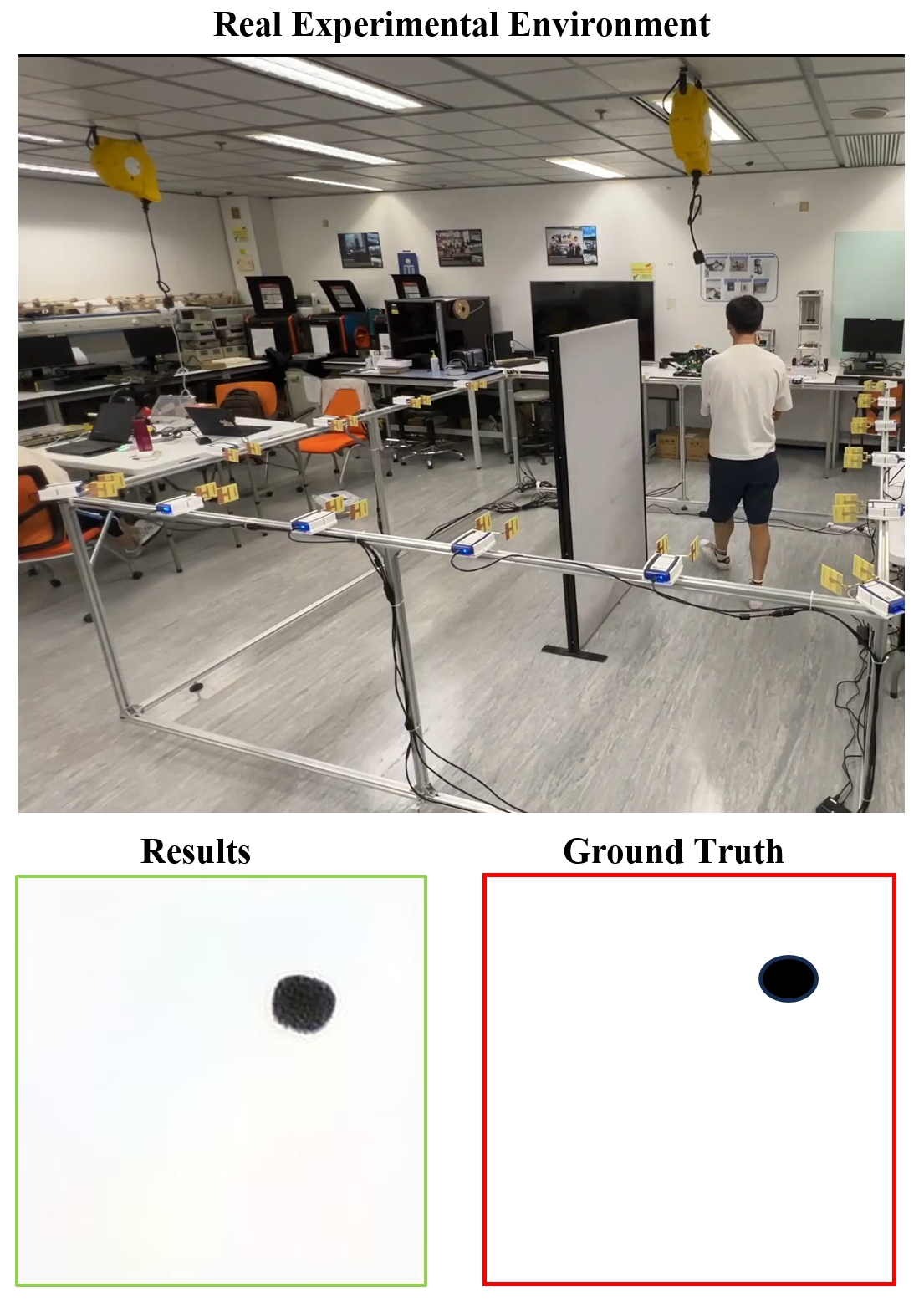}
	\caption{WiFi-GEN results for human perception in real environments. (Ground truth is a simplified 2D human body with a known center point.)}
	\label{fig:pe}
\end{figure}

To validate robustness, we conducted a laboratory experiment closely matching the simulated scenario. 
A $3 \times 3$ m$^2$ DoI was deployed with 20 ADALM-PLUTO software-defined radios evenly placed along the boundary, similar to conventional setups \cite{deshmukh2022physics} but with fewer WiFi nodes. 
As shown in Fig.~\ref{fig:pe}, WiFi-GEN successfully interprets real-world measurements, accurately capturing object shape and spatial location despite variations in signal distribution. 
Notably, it reconstructs the human body’s shape and position with high fidelity.  

\section{Conclusions and Future Work}\label{conclusion}
This work demonstrates the feasibility of applying generative AI to WiFi-based indoor imaging, a domain traditionally dominated by physics-based wave equation solutions. 
By addressing challenges of ill-posedness, non-linearity, and uncertainty, WiFi-GEN achieves about threefold higher shape reconstruction accuracy, higher resolution, and larger DoI compared with physical methods. 
Laboratory experiments further confirm that a pre-trained model generalizes well to real environments with different signal conditions. 
Future work will explore scaling to larger environments, handling richer object categories, and integrating multimodal sensing.

\section{Acknoledgement}
This work was partially supported by the National Key R\&D Program of China (Grant Nos. 2024YFE0200801 and 2024YFE0200804), Natural Science Foundation of Top Talent of SZTU (grant no. GDRC202320), Project for Improving Scientific Research Capabilities of Key Construction Disciplines in Guangdong Province (2025ZDJS039), and Shenzhen Science and Technology Program (No. JCYJ2024081311 3218025). The
work of R. Murch was supported by the Hong Kong Research Grants Council
Area of Excellence Grant AoE/E-601/22-R.

\bibliographystyle{IEEEbib}
\bibliography{strings,refs}

\end{document}